%% file: paper.tex
\documentclass[]{bytedance_seed}

\input{macros}

\usepackage{marvosym}

\newcommand{\myparagraph}[1]{\vspace{0.1em}\noindent\textbf{#1}}

\title{
    Translation as a Bridging Action: Transferring Manipulation Skills from Humans to Robots
}

\author[1,2,*]{Sijin Chen}
\author[2,*]{Kaixuan Jiang}
\author[2,\dagger,\text{\Letter}]{Haixin Shi}
\author[2]{Yanhui Wang}
\author[2]{Weiheng Zhong}
\author[2]{Haosheng Li}
\author[2]{Bo Jiang}
\author[2]{Yuxiao Liu}
\author[1,\text{\Letter}]{Xihui Liu}

\affiliation[1]{HKU-MMLab}
\affiliation[2]{ByteDance Seed}

\contribution[]{$^*$Equal Contribution, $^\dagger$Project Lead, \textsuperscript{\Letter}Corresponding Author}

\abstract{
    We study whether we can learn novel manipulation skills from human actions to a bi-manual robot with parallel grippers.
    Human action data is cheap, abundant, and diverse, making it one of the most promising resources for scaling up robot learning.
    Yet transferring skills from humans to robots remains hard: most prior work treats humans as just another bi-manual 6DoF embodiment, where hand-pose estimates are noisy and the contact patterns of human fingers differ fundamentally from those of a parallel gripper.
    We argue that learning rotation-inclusive action signals from human data is therefore sub-optimal, and instead propose a bridging action representation: the relative wrist translation within the initial head-camera frame, an action space shared by humans and robots.
    To handle the potential absence of certain action components in different embodiments, we build a $\pi_0$-like vision-language-action model with interleaved action tokens and attention masking.
    On a suite of novel bi-manual manipulation tasks, our bridging action transfers human manipulation knowledge to robots far more effectively than noisy 6DoF human actions and scales with the amount of human data.
}

\date{\today}
\correspondence{
    Xihui Liu at \email{xihuiliu@eee.hku.hk}, 
    and
    Haixin Shi at \email{shihaixin@bytedance.com}
}

\checkdata[Project Page]{\url{https://translation-as-a-bridging-action.github.io/}}

\begin{document}
\maketitle

\begin{figure}[h]
    \centering
    \makebox[\textwidth][c]{
        \includegraphics[width=\textwidth]{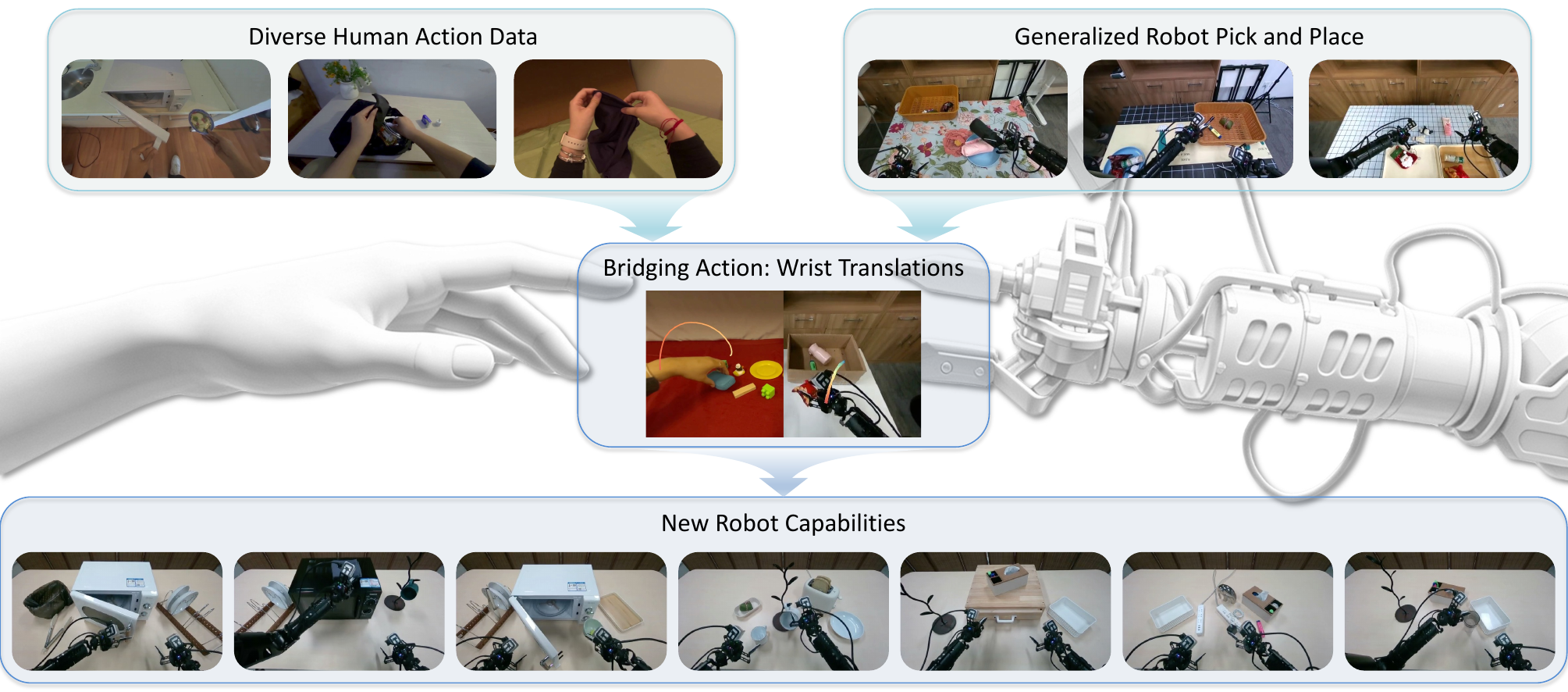}
    }
    \caption{
        \textbf{Overview.}
        We study whether we can transfer human manipulation skills to bi-manual robots with grippers by learning relative wrist translation as the bridging action representations.
        We show that such bridging action not only enables robust and efficient manipulation knowledge transfer, but can benefit from large-scale pre-training.
    }
    \label{fig:teaser}
\end{figure}

\input{sections/intro}
\input{sections/related}
\input{sections/method}

\input{sections/exp}

\input{sections/conclusion}

\clearpage

\bibliographystyle{plainnat}
\bibliography{citation}

\end{document}

%% file: macros.tex
\usepackage{minitoc}
\usepackage{url}
\usepackage{amssymb}
\usepackage[utf8]{inputenc}
\usepackage{microtype}
\usepackage{booktabs}
\usepackage{pifont} 
\usepackage{multirow}
\usepackage{makecell}
\usepackage{paralist}
\usepackage{xspace}
\usepackage{color}
\usepackage{xcolor}
\usepackage{colortbl}
\usepackage{adjustbox}
\usepackage{hyperref} 
\usepackage[edges]{forest}
\usepackage{tikz} 
\usepackage{caption}
\usepackage{amsfonts}
\usepackage[toc,page,header]{appendix}
\usepackage[utf8]{inputenc}
\usepackage{bbm}
\usepackage{enumitem}

\definecolor{seedc}{RGB}{7, 92, 173}

\newcommand{\name}[1]{GR-3}

\newcommand{\hardware}[1]{ByteMini}

\renewcommand{\paragraph}[1]{\vspace{0.1em}\noindent\textbf{#1}}

%% file: sections/intro.tex
\section{Introduction}

In this paper, we study whether we can learn novel manipulation skills from human data on a bi-manual robotic platform.
Benefiting from the robot-free setting, the collection of human manipulation data can be conducted at a much lower cost and almost anywhere~\cite{generalist2025gen0,grauman2022ego4d}.
Therefore, in terms of environmental and skill diversity, human data is considered one of the most scalable, economical, and promising ways to scale up robot learning~\cite{hoque2025egodex,generalist2025gen0,zheng2026egoscale}.
Along with the growing popularity of ego-centric human data collection~\cite{grauman2022ego4d,hoque2025egodex,grauman2024ego}, researchers have flocked to the exploration of how human data can improve embodied policies.

Prior research has proposed various types of human action representations for ego-centric human data.
A large proportion of work adopts wrist poses~\cite{kareer2025egomimic,zhu2026emma,punamiya2025egobridge,cheang2025gr3,kareer2025emergence} with optional fingertip positions~\cite{cai2025n,qiu2025humanoid}, parameterized hands~\cite{romero2022embodied,luo2025being,luo2026being,yang2025egovla}, or finger joints~\cite{zheng2026egoscale} as human actions.
In addition to explicit action representation, there are also attempts to learn latent action representations from ego-centric human videos~\cite{chen2025moto,fan2025xr,bruce2024genie,ye2024latent}.
Incorporating human actions when building robot policies has shown great potential for improving generalization~\cite{kareer2025egomimic,punamiya2025egobridge,cai2025n}, reducing the reliance on robot data~\cite{zheng2026egoscale,bi2026h}, and steering pre-trained policies to manipulate novel objects and follow new instructions~\cite{cheang2025gr3,kareer2025emergence,yuan2025motiontrans} via human-robot co-training.

Different from cross-embodiment robot data~\cite{o2024open,wu2024robomind,hou2025robomind,wu2025robocoin}, learning from human data encounters unique challenges. 
Human actions are typically derived from hand pose estimators~\cite{kareer2025emergence,cheang2025gr3,kareer2025egomimic,yang2025egovla}.
Though human hands can be constrained by parameterization~\cite{luo2025being,luo2026being,pavlakos2024reconstructing,romero2022embodied,yang2025egovla} or re-targeting~\cite{zheng2026egoscale}, the extracted actions are inevitably noisy.
Additionally, contact patterns differ greatly between human hands and robot grippers~\cite{wei2026one}: the additional degrees of freedom in fingers make wrist rotations less meaningful when reflecting manipulation behaviors.
Therefore, we argue that \emph{learning rotation-inclusive robotic manipulation signals from human data is highly challenging and sub-optimal}.
Instead, we propose to learn only the wrist translations as a shared action space between humans and robots.
Since humans and robots both act on what they perceive, we extract the relative wrist translations under the head camera frame.
To this end, we end up with two different action spaces: 1) the rotation-inclusive end-effector action to control the robot~\cite{zheng2026egoscale,cheang2025gr3,kareer2025emergence} and 2) the wrist-translation action that can be shared between humans and robots.

To enable efficient manipulation knowledge transfer from human to robot, we are required to design a unified training objective for a multi-embodiment model.
We build our method upon a $\pi_0$-like~\cite{black2024pi_0} vision-language-action model, which processes the observation and instruction inputs with pre-trained vision-language priors and generates actions via flow matching~\cite{liu2022flow,lipman2022flow}.
Since we are required to handle the potential missing action components across different embodiments, we propose to adopt interleaved action tokens to model wrist translation, 6DoF end-effector action, and gripper signals.
By masking the missing action components within the attention layers, we can handle actions from different embodiments via variable length inputs.
Critically, we find that randomly adding and substituting wrist translations for 6DoF end-effector actions on robot action data is essential for transferring wrist translation knowledge to executable robot end-effector actions.

We adopt a three-stage training strategy by first pre-training the model on a large-scale collection of human action data with only wrist translation actions.
Then, we co-train the model by incorporating generalized pick-and-place robot actions~\cite{cheang2025gr3} with task-specific human actions (\textit{e.g.}, opening microwaves, wiping, stacking, \textit{etc.}).
Empirically, we find that the robot is able to complete certain tasks without task-specific robot demonstrations.
We can also improve the overall performance of the model by post-training on an additional set of few-shot real-robot data.
To conclude, our contributions include:

\begin{itemize}
    \item We propose a bridging action representation based on wrist translations, which is shared between robots and humans and is robust to noisy human actions.
    We show that the bridging action facilitates superior manipulation skill transfer from human to robot than simply adopting noisy 6DoF human actions.

    \item To cope with the heterogeneous action components across data sources, we design a unified, interleaved action representation that handles missing action components through attention masks.
    This interleaved design also enables the model to benefit from large-scale human action pre-training.

    \item We further characterize the upper bound of our bridging representation in robotic skill transfer, suggesting its potential to scale up to broader cross-embodiment learning settings.
\end{itemize}

%% file: sections/related.tex
\section{Related Works}

\myparagraph{Learning from human.}
Along with the growing number and size of ego-centric datasets~\cite{zhou2018towards,goyal2017something,miech2019howto100m,damen2018scaling,grauman2022ego4d,grauman2024ego,hoque2025egodex}, researchers have explored diverse approaches to harness physical knowledge from large-scale human data.
Some works~\cite{cheang2024gr,wu2023unleashing,hu2024video,li2026causal} learn generative visual representations from human videos.
Another line of work learns intermediate representations from human videos, \textit{e.g.}, affordance~\cite{bahl2023affordances,bahl2022human}, keypoints~\cite{wang2023mimicplay,wen2023any,bharadhwaj2024track2act}, and latent actions~\cite{bruce2024genie,ye2024latent,bu2025agibot,chen2025moto,fan2025xr,bjorck2025gr00t}.
There are also works that collect explicit human actions directly with gloves~\cite{xu2025dexumi,tao2025dexwild} or extract human actions from videos using hand pose estimation~\cite{kareer2025emergence,kareer2025egomimic,punamiya2025egobridge,zhu2026emma,luo2026being,luo2025being}.
Human data, a cost-effective manipulation data source, is not only regarded as the pretraining data source~\cite{zheng2026egoscale,cheang2024gr,bjorck2025gr00t}, but is also widely adopted during human-robot alignment~\cite{kareer2025egomimic,punamiya2025egobridge,zhu2026emma,shi2026egohumanoid} and even post-training~\cite{ye2026world,cheang2025gr3,kareer2025emergence} to improve robot foundation models~\cite{black2024pi_0,intelligence2025pi_0.5,ye2026world}.
In our project, we learn explicit human actions and argue that wrist rotations are unreliable on noisy human data and semantically misaligned with parallel grippers.
Instead, we establish a translation-based bridging action representation shared between humans and robots to enable effective human to robot manipulation knowledge transfer.

\myparagraph{Cross embodiment learning} aims to learn the shared knowledge across different robot configurations within a unified model to control a diverse collection of robots~\cite{o2024open,team2024octo,black2024pi_0,intelligence2025pi_0.5,yang2024pushing}.
Learning from explicit human actions can also be considered a type of cross embodiment learning~\cite{kareer2025egomimic,punamiya2025egobridge,zheng2026egoscale,zhu2026emma}.
However, the differences among action representations pose significant challenges.
Some works define a unified action space by concatenating~\cite{liu2024rdt,cheang2025gr3,kareer2025egomimic}, padding missing dimensions~\cite{black2024pi_0,pertsch2025fast,intelligence2025pi_0.5}, or simply building different projectors~\cite{wang2024scaling,bjorck2025gr00t} for different action representations.
In this project, we co-train human and robot actions via a shared bridging action representation and adopt an interleaved action sequence to handle the potential missing action components within different data sources.
We show that our proposed action representation and training strategy enable effective human to robot manipulation knowledge transfer.

\myparagraph{Data pyramid and co-training} are widely discussed when developing robust and generalizable robot learning systems~\cite{bjorck2025gr00t,barreiros2025careful,lin2026systematic,generalist2026gen1,cheang2025gr3}.
Instead of resorting to the costly practice of collecting real robot data, cross-embodiment data~\cite{o2024open,kim2024openvla,liu2024rdt,black2024pi_0,intelligence2025pi_0.5}, human data~\cite{kareer2025egomimic,zheng2026egoscale,cheang2025gr3}, vision-language corpora~\cite{goyal2017making,chen2015microsoft,li2024llava,zitkovich2023rt2,wu2026pragmatic}, and videos~\cite{bu2025agibot,bjorck2025gr00t,cheang2024gr,wu2023unleashing,hu2024video,ye2026world,li2026causal} serve as more cost-effective alternatives for scaling up robot learning.
By exposing the model to a more diverse data source and supervision, the robot is more robust to environmental distortions~\cite{cheang2025gr3,intelligence2025pi_0.5,lin2026systematic} and can follow more generalized instructions~\cite{zitkovich2023rt2,cheang2025gr3}.
In this paper, our work focuses specifically on how to transfer human manipulation skills to robots and whether it can benefit from large-scale human action pre-training.

%% file: sections/method.tex
\section{Hardware and Platform}
\label{sec:hardware}

\myparagraph{Robot platform.}
We experiment with the ByteMini robot~\cite{cheang2025gr3} (Fig.~\ref{exp:fig:eval_setup}.b), which is a bi-manual mobile manipulation platform with two 7-DoF arms, parallel grippers, and three RGB-D cameras mounted on the head and two wrists.
During data collection, we ask the tele-operators to randomly adjust the robot's height, position, and rotation for diversity.
While during rollout, the robot base is kept static, and the arms are controlled with 6DoF end-effector poses and discrete gripper signals.

\myparagraph{Human data platform.}
We collect our in-lab human action data with PICO 4 Ultra Enterprise.
Since we are mainly interested in learning transferable robotic manipulation behaviors from human data, we ask the operators to imitate the robot gripper with their hand postures and to keep their hands within the top camera's field of view during human action data collection.

\begin{figure}[t]
    \centering
    \includegraphics[width=1\linewidth]{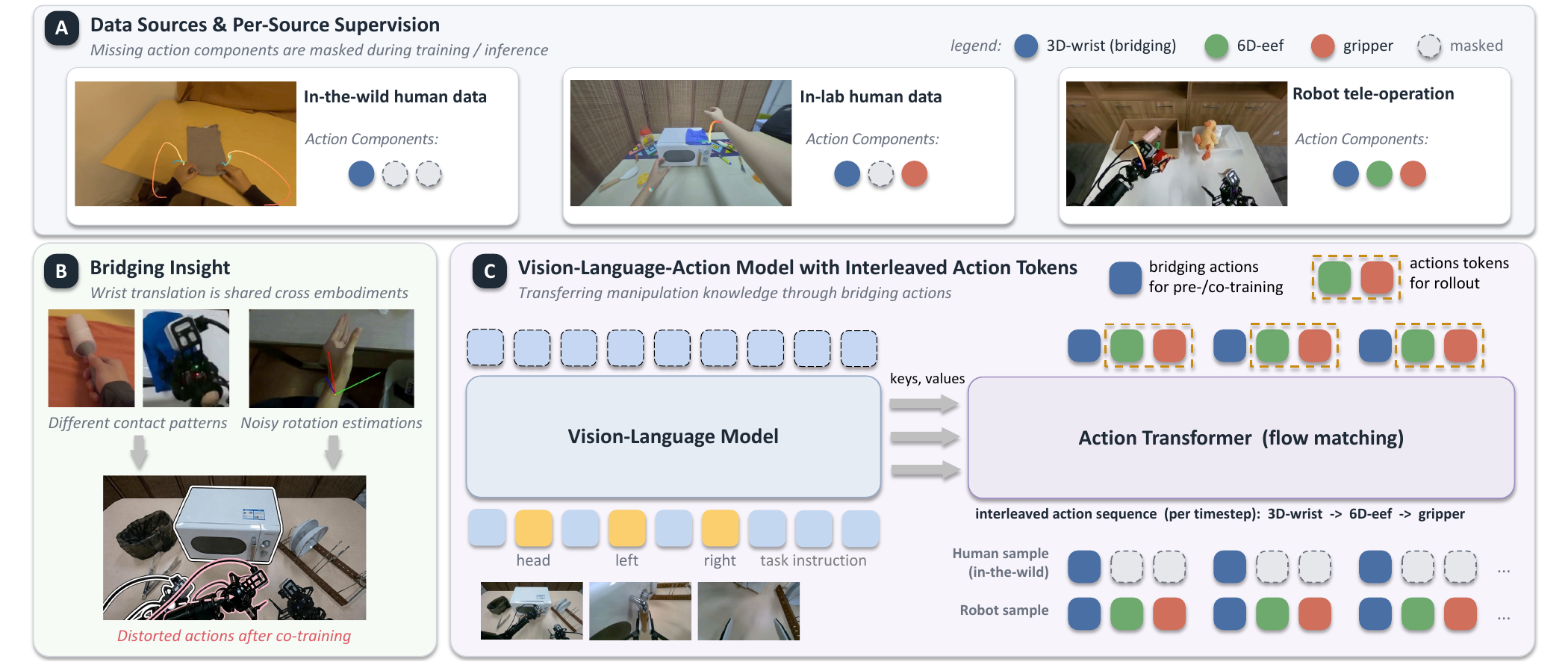}
    \caption{
        \textbf{Architecture Overview.}
        We train our model on a) the mixture of human and robot action data.
        As shown in b), learning from 6DoF human action is challenging and suboptimal because of the difference in contact patterns and noisy hand pose estimations.
        We adopt c) a $\pi_0$-like~\cite{black2024pi_0} vision-language-action model as our base policy and adopt interleaved action sequences to handle potential missing action components and enable manipulation behavior transfer.
    }
    \label{method:fig:pipeline}
\end{figure}

\section{Method}
\label{sec:method}

We propose a framework that can transfer manipulation skills from human actions to a bi-manual robot.
Instead of directly using 6DoF wrist actions, we propose to adopt a bridging action representation based on wrist translations.
Such bridging action can be extracted from both human and robot actions.
To handle the potential missing action components for different action data sources, we build an interleaved action sequence for flow matching.

\subsection{Motion Bridging Action Representation}
\label{sec:method:action_repr}

\myparagraph{Why not just 6DoF wrist actions for human actions?}
The mainstream practice treats humans as another embodiment by extracting relative 6DoF wrist actions from hand-pose estimators~\cite{kareer2025egomimic,kareer2025emergence,cheang2025gr3,yuan2025motiontrans,zheng2026egoscale}.
However, we argue this is sub-optimal: 1) the wrist rotations for human data are noisy due to predictor errors~\cite{wei2026one}, and 2) the distinct contact patterns between fingers and parallel grippers decouple wrist rotation from the semantic manipulation behavior. 
Empirically, we find that directly replaying the extracted human 6DoF wrist actions on robots often leads to distorted behaviors.

\myparagraph{From wrist poses to a shared bridging signal $\mathbf{a}^{\text{3D-wrist}}$.}
Our key insight is that both humans and robots act upon what they perceive, therefore, we treat \emph{the relative wrist translation observed from the head camera} as the bridging action.
We first project the wrist pose into the head-camera frame.
Let $\mathbf{W}^{t}_{w}\!\in\!\mathbb{SE}(3)$ denote the wrist pose in the world frame at time $t$, and $\mathbf{T}^{t}_{w\leftarrow c}\!\in\!\mathbb{SE}(3)$ the head-camera pose at time $t$, whose camera frame we abbreviate as $\mathrm{c}_t$.
Applying $(\mathbf{T}^{t}_{w\leftarrow c})^{-1}$ maps the wrist pose into $\mathrm{c}_t$, yielding $\mathbf{W}^{t+i}_{\mathrm{c}_t} = (\mathbf{T}^{t}_{w\leftarrow c})^{-1}\,\mathbf{W}^{t+i}_{w}$. 
We take the translation component as the relative wrist translation over a $k$-step future window:
\begin{equation}
\boldsymbol{a}^{\text{3D-wrist}}_{t+i}
= \Delta \mathbf{W}^{\text{3D}}
= \boldsymbol{t}\!\left( \mathbf{W}^{t+i}_{\mathrm{c}_t} \right)
- \boldsymbol{t}\!\left( \mathbf{W}_{\mathrm{c}_t}^{t} \right), \qquad i = 1, \dots, k,
\end{equation}
where $\boldsymbol{t}(\cdot)$ extracts the $3\!\times\!1$ translation components in $\mathbb{SE}(3)$. 
In practice, we concatenate wrist translations for both arms, yielding $\mathbf{a}_{t}^{\text{3D-wrist}}\!\in\!\mathbb{R}^{k\times 6}$  for both bi-manual robots and humans.
Such translation-only bridging action $\mathbf{a}_{t}$ serves as the bridge between the two embodiments.
To sum up, $\mathbf{a}_{t}^{\text{3D-wrist}}$ is 1) physically meaningful for describing motion under a shared observation perspective, 2) robust to noisy rotation estimates, and 3) embodiment-agnostic by construction.

\myparagraph{Robot end-effector action $\mathbf{a}^{\text{6D-eef}}$.}
The robot 6DoF end-effector action is defined as the relative wrist motion with respect to the initial end-effector pose: 
\begin{equation}
\boldsymbol{a}^{\text{6D-eef}}_{t+i}
= \Delta \mathbf{W}^{\text{6D}} =  \left( \mathbf{W}^{t}_{w} \right)^{-1} \mathbf{W}^{t+i}_{w},
\qquad i = 1, \dots, k.
\end{equation}
Being a relative pose between two $\mathbb{SE}(3)$ elements, this action representation is invariant to the absolute camera pose and is physically meaningful for describing arm motions~\cite{zheng2026egoscale}.
We further transform $\mathbb{SE}(3)$ into Cartesian coordinates and Euler angles for robot actions, resulting in $\mathbf{a}_{t}^{\text{6D-eef}}\!\in\!\mathbb{R}^{k\times 12}$ by concatenating the end-effector actions for both arms.

\myparagraph{Gripper action $\mathbf{a}^{\text{gripper}}$}
is defined as a chunk of binary signal $a^{\text{gripper}}_i \in \{0, 1\}$ per gripper, where we use $1$ for close and $0$ for open.
We can also annotate hand closure as gripper signal for our in-lab human data.
The gripper action is defined as $\mathbf{a}^{\text{gripper}}_t\!\in \!\mathbb{R}^{k\times 2}$ for humans and bi-manual robots.

\myparagraph{Unified action and per-embodiment supervision.}
We define a unified action space via the concatenation of different action components 
$\mathbf{a}_t = (\mathbf{a}^{\text{3D-wrist}}_t, \mathbf{a}^{\text{6D-eef}}_t, \mathbf{a}^{\text{gripper}}_t)$. 
Different data sources contain different subsets of the action components, and we only supervise the corresponding components that are reliably available.
As shown in Table~\ref{method:tab:supervision}, we learn the shared $\mathbf{a}^{\text{3D-wrist}}$ for human actions while additionally grounding $\mathbf{a}^{\text{3D-wrist}}$ into the executable $\mathbf{a}^{\text{6D-eef}}$ for robot actions.

\begin{table}[htbp]
\caption{
    \textbf{Per-embodiment action supervision.}
    Based on reliably extractable action components, we learn distinct action representations for various embodiments under a unified action space.
}
\label{method:tab:supervision}
\centering
\small
\begin{tabular}{l|ccc}
\toprule
Data source & $\mathbf{a}^{\text{3D-wrist}}$ & $\mathbf{a}^{\text{6D-eef}}$ & $\mathbf{a}^{\text{gripper}}$ \\
\midrule
In-the-wild human action data (EgoDex~\cite{hoque2025egodex} + out-sourced) & $\checkmark$ & --           & --           \\
In-lab human action data & $\checkmark$ & --           & $\checkmark$ \\
Robot tele-operation                     & $\checkmark$ & $\checkmark$ & $\checkmark$ \\
\bottomrule
\end{tabular}
\end{table}

\subsection{VLA Model with Interleaved Action Sequence}
\label{sec:model:model}

\myparagraph{Architecture.}
Our model is a $\pi_0$-like~\cite{black2024pi_0,cheang2025gr3} end-to-end VLA, $\pi_\theta(l, o_t)$, which generates an action chunk $\mathbf{a}_t = a_{t:t+k}$ conditioned on the language instruction $l$ and the observations $o_t$ from the head and two wrist cameras.
We pad blank images for the missing wrist views in human action data.

The VLA model formulates vision, language, and action tokens within a sequence, sharing the self-attention layer, but with two separate sets of parameters to balance different training objectives.
The vision and language tokens ($o_t$, $l$) are processed via a pre-trained VLM~\cite{bai2025qwen25vltechnicalreport}, and the resulting vision-language KV-cache serves as the context condition for the Action Transformer to generate the action chunk via flow matching~\cite{liu2022flow,lipman2022flow}.

\myparagraph{Interleaved action tokens.}
Given previous analysis on action definitions in Sec.~\ref{sec:method:action_repr}, an action chunk can be represented with a subset of $\{\mathbf{a}^{\text{3D-wrist}},\mathbf{a}^{\text{6D-eef}},\mathbf{a}^{\text{gripper}}\}$ depending on the data source. 
We exploit the Transformer's ability to handle variable-length inputs via attention masks and position ids~\cite{vaswani2017attention}, and organize the action tokens in the
$\mathbf{a}^{\text{3D-wrist}} \rightarrow \mathbf{a}^{\text{6D-eef}} \rightarrow \mathbf{a}^{\text{gripper}}$ order to handle possible absence of certain action types.
We build corresponding input/output layers for different action tokens, while sharing the action transformer.

The ordering of action components is established based on two priors: 
1) the shared bridging signal should be attended to by the 6DoF action tokens, enabling explicit manipulation knowledge transfer from human to robot within the attention pattern itself; 
and 2) the gripper signal is typically triggered after the end-effector reaches its target.
When handling missing action components, it is natural to mask certain action tokens in attention layers and omit the loss calculation, \textit{e.g.}, the 6DoF end-effector actions for human data.

\myparagraph{Flow-matching action objective.}
The action generation is trained with flow matching~\cite{lipman2022flow,liu2022flow}. 
Given $\tau\!\in\!(0,1)$ and $\epsilon\!\sim\!\mathcal{N}(\mathbf{0}, \mathbf{I})$, the model is required to denoise the noisy action chunk $\mathbf{a}^{\tau}_{t} = \tau\,\epsilon + (1-\tau)\,\mathbf{a}_{t}$ based on observation $o_t$ and language $l$ inputs, by predicting the velocity $\hat v(\mathbf{a}^{\tau}_t, o_t, l, \tau)$ from the noise $\epsilon$ towards the ground-truth action $\mathbf{a}_{t}$.
The ground-truth velocity is defined as $v^* = \epsilon - \mathbf{a}_{t}$, and the action flow matching loss is defined as:
\begin{equation}
    \mathcal{L}_{\text{FM}} = \left\|\hat v(\mathbf{a}^{\tau}_{t}, o_t, l, \tau) - v^*\right\|_2^2.
\end{equation}
For each training sample, $\mathcal{L}_{\text{FM}}$ is computed on the provided action components. 
During inference, we generate only $\mathbf{a}^{\text{6D-eef}}$ and $\mathbf{a}^{\text{gripper}}$ for robot control by integrating the velocity from $\tau\!=\!0$ to $1$ with $\Delta \tau = 0.2$ using the Euler method, \textit{i.e.}, $\mathbf{a}^{\tau + \Delta \tau}_{t} = \mathbf{a}^{\tau}_{t} + \Delta\tau \cdot \hat v(\mathbf{a}^{\tau}_{t}, o_t, l, \tau)$, across all the experiments.

\myparagraph{Vision-language co-training.}
To prevent over-fitting on action data, we co-train the VLA model with a collection of vision-language data~\cite{cheang2025gr3,intelligence2025pi_0.5,zitkovich2023rt2} using the standard next-token prediction (NTP) objective, \textit{i.e.}, 
$\mathcal{L}_{\text{NTP}} = - \frac{1}{|s|}\sum\nolimits_{i=1}^{|s|} \log P\!\left(s^{i} \,\middle\vert\, s^{[1, \dots, i-1]};\, o_{t}, l\right)$,
where $s$ is the response sequence, and $|s|$ is the length of $s$.
Each training sample adopts either $\mathcal{L}_{\text{FM}}$ or $\mathcal{L}_{\text{NTP}}$ for action data or vision-language data, respectively.
Therefore, the total loss can also be represented as a weighted sum over the batch.

\subsection{Training Strategies}
\label{sec:method:strategy}

\myparagraph{Stage I: Pre-training on human actions.}
Human data offers unmatched skill and scene diversity, is independent of robot hardware, and is one of the most scalable manipulation data sources~\cite{hoque2025egodex,grauman2022ego4d,zheng2026egoscale}.
Accordingly, we pre-train our VLA model on approximately 600 hours of human actions: covering $\sim$70 hours from selected tasks in EgoDex~\cite{hoque2025egodex}, $\sim$500 hours of out-sourced free-form household manipulation, and $\sim$45 hours of in-lab human actions.
The out-sourced and in-lab human actions are both collected with PICO 4 Ultra Enterprise~\cite{cheang2025gr3}.
Because the contact pattern and the rotation estimation cannot be reliably standardized at this scale, we supervise only the bridging signal $\mathcal{L}_{\text{FM}}^{\text{3D-wrist}}$ for pre-training.

\myparagraph{Stage II: Human-robot co-training.}
We now need the model to learn to generate executable robot actions by exposing the model to real robot trajectories.
We adopt an aggressive data strategy by using only \textbf{generalized robot pick-and-place data} ($\sim$72 hours) over 100 objects, annotated with the fixed prompting format ``\texttt{put \{object\} into \{container\}}''~\cite{cheang2025gr3} to distinguish the robot actions from the skills in human actions.
Paired with robot data, we collect $\sim$3 hours per task of task-specific \textbf{in-lab human actions} across 15 tasks, e.g., ``\texttt{open the microwave door}'', \textit{etc.}
To minimize contact-pattern mismatch, we ask in-lab operators to mimic the gripper with their hand posture, and we annotate hand closure as the gripper control signal.
All three losses, $\mathcal{L}_{\text{FM}}^{\text{3D-wrist}}$, $\mathcal{L}_{\text{FM}}^{\text{6D-eef}}$, and $\mathcal{L}_{\text{FM}}^{\text{gripper}}$ are active on robot data.
In practice, we randomly include $\mathbf{a}_t^{\text{3D-wrist}}$ or substitute $\mathbf{a}_t^{\text{3D-wrist}}$ for $\mathbf{a}_t^{\text{6D-eef}}$ as the prediction target to explicitly bind the bridging representation to the executable robot actions. 
We show that this binding strategy is essential for learning transferable manipulation skills in Sec.~\ref{sec:exp:bridging_tasks}.

\myparagraph{Stage III: Few-shot robot post-training.}
We collect an additional set of robot data, containing 100 tele-operated robot actions per task in our in-lab human data.
For post-training, we only use 10 trajectories per task at this stage to study the data efficiency (Sec.~\ref{exp:sec:few-shot}).

\myparagraph{Implementation.}
Our VLA model adopts the Mixture-of-Transformer~\cite{liang2024mixture,black2024pi_0} architecture with approximately 4B parameters.
In \textbf{Stage I} (human-only pre-training), we initialize the VLA model from a pre-trained VLM~\cite{bai2025qwen25vltechnicalreport} and train all parameters with a global batch size of 1024 for 400k iterations.
In \textbf{Stage II} (human--robot co-training), we continue to co-train all the parameters on both robot and human trajectories with a global batch size of 256 for another 120k iterations.
In \textbf{Stage III} (few-shot robot post-training), we further fine-tune the model by co-training on additional few-shot robot trajectories, also with a global batch size of 256 for another 25k iterations.
Additionally, we follow prior practice~\cite{li2024autoregressive,cheang2025gr3} to increase the action transformer's effective batch size by repeating the VLM's KV-cache by $4\times$ for each training batch to accelerate convergence.

%% file: sections/exp.tex
\section{Experiments}
\label{sec:exp}

We conduct real-world experiments to study how the bridging action representation helps transfer manipulation skills from human actions to robots.
Concretely, we aim to answer the following questions:

\begin{itemize}[leftmargin=1.5em,itemsep=0.1em]
    \item \textbf{Q1.} Does the bridging action transfer manipulation skills beyond generic pick-and-place, and does it scale with large-scale human-only pre-training? (Sec.~\ref{exp:sec:principle})
    \item \textbf{Q2.} How is the bridging action when comparing to 6DoF human actions? (Sec.~\ref{exp:sec:vs6dof})
    \item \textbf{Q3.} Does human-only pre-training improve the data efficiency of few-shot robot post-training? (Sec.~\ref{exp:sec:few-shot})
    \item \textbf{Q4.} What training objective is essential during human-robot co-training? (Sec.~\ref{sec:exp:bridging_tasks})
    \item \textbf{Q5.} Does human-only pre-training objective align with the executable robot action space? (Sec.~\ref{exp:sec:alignment})
    \item \textbf{Q6.} What is the performance upper bound of the bridging action? (Sec.~\ref{exp:sec:upperbound})
\end{itemize}

\subsection{Evaluation Setups}
\label{exp:sec:eval_setup}

\begin{figure}[t]
    \centering
    \begin{subfigure}[b]{1\textwidth}
        \centering
        \includegraphics[width=\textwidth]{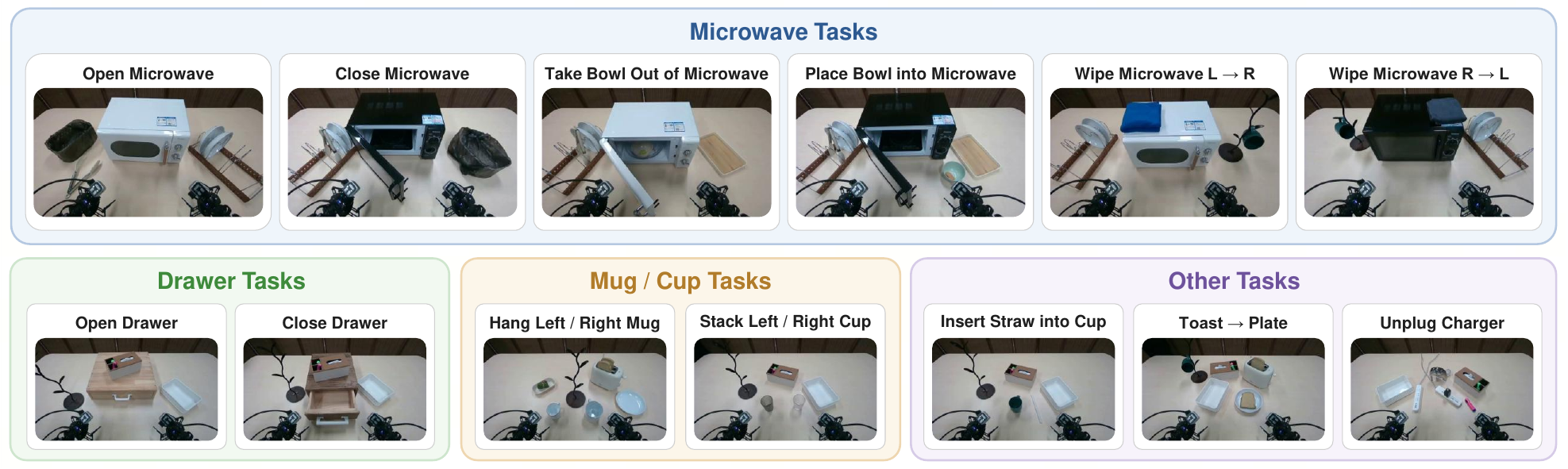}
        \label{fig:sub1}
    \end{subfigure}
    \begin{subfigure}[b]{1\textwidth}
        \centering
        \includegraphics[width=\textwidth]{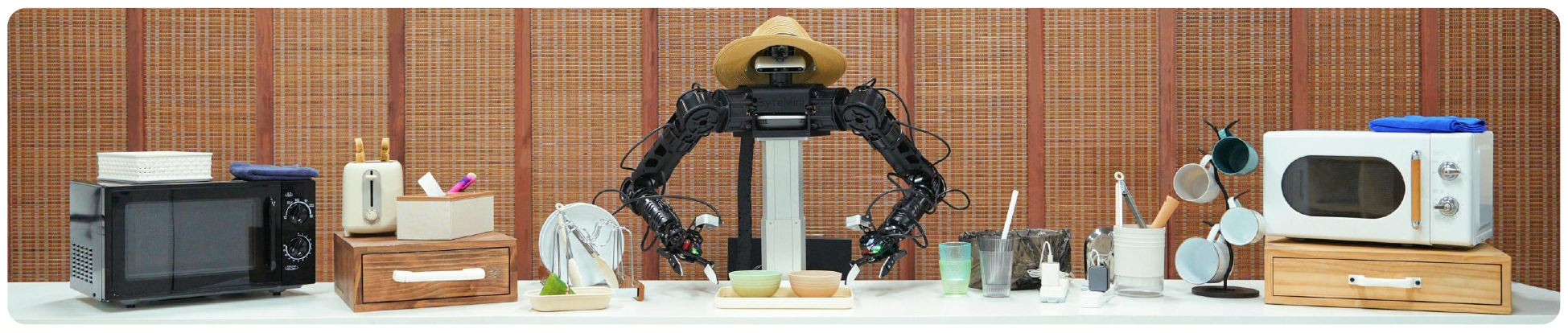}
        \label{fig:sub2}
    \end{subfigure}
    \caption{
        \textbf{Evaluation setups.} 
        We show one of the initialized setups for each task (top), and the objects used throughout policy evaluations (bottom).
    }
    \label{exp:fig:eval_setup}
\end{figure}

\begin{figure}[htbp]
    \centering
    \includegraphics[width=1\linewidth]{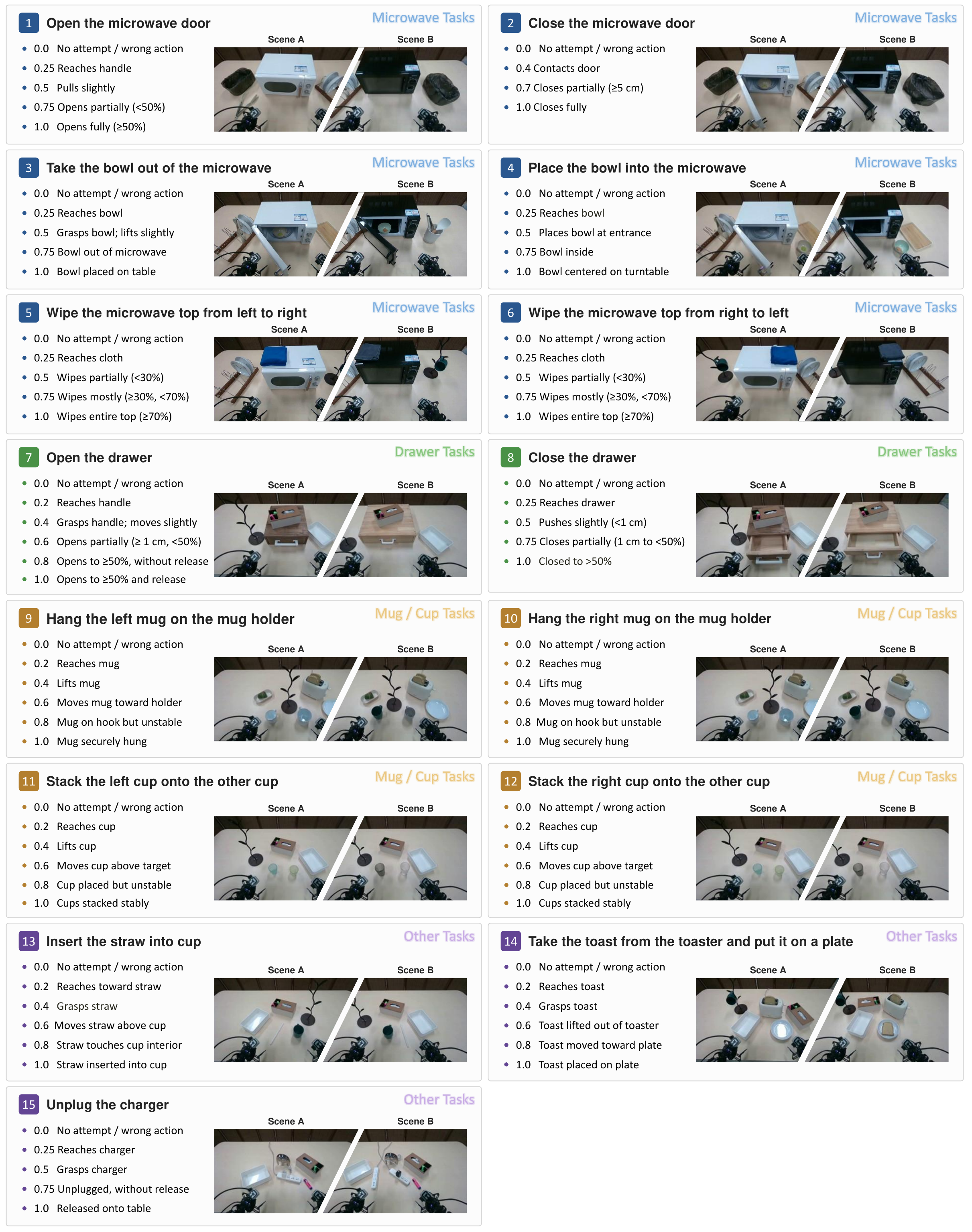}
    \caption{
        \textbf{Overview of real-world evaluation tasks.} 
        We experiment with 15 manipulation tasks, categorized with respect to their manipulation objects.
        For each task, we adopt two distinct testing layouts and a detailed progress scoring criterion.
    }
    \label{exp:fig:appdxB_test}
\end{figure}

\myparagraph{Tasks.}
We evaluate our policy on 15 manipulation tasks as shown in Fig.~\ref{exp:fig:eval_setup}, which is grouped by their manipulation objects (microwave, drawer, mug/cup, and others).
We also provide the task progress evaluation criteria in Fig.~\ref{exp:fig:appdxB_test}.
For each of listed tasks, we collect 1) three hours of task-specific in-lab human actions, and 2) 100 task-specific in-lab robot trajectories.

\begin{enumerate}[leftmargin=1.6em,itemsep=0.1em]
    \item \textbf{Open the microwave door.} The robot reaches for the microwave door handle, grasps it, and pulls it outward till the door is opened.
    \item \textbf{Close the microwave door.} The robot pushes the microwave door until it is fully closed.
    \item \textbf{Take the bowl out of the microwave.} The robot reaches inside the open microwave, picks the bowl up, and places it on the table while avoiding collisions.
    \item \textbf{Place the bowl into the microwave.} The robot picks up the bowl from the table and places it inside the open microwave while avoiding collisions.
    \item \textbf{Wipe the microwave top from left to right.} The robot grasps the towel on the left top of the microwave and pushes it towards the right top.
    \item \textbf{Wipe the microwave top from right to left.} The robot grasps the towel on the right top of the microwave and pushes it towards the left top.
    \item \textbf{Open the drawer.} The robot reaches for the drawer handle, grasps it, and pulls it outward.
    \item \textbf{Close the drawer.} The robot pushes the drawer until it is fully closed.
    \item \textbf{Hang the left mug on the mug holder.} The robot picks up the left mug on the table and hangs it onto the corresponding side of the mug holder.
    \item \textbf{Hang the right mug on the mug holder.} The robot picks up the right mug on the table and hangs it onto the corresponding side of the mug holder.
    \item \textbf{Stack the left cup onto the other cup.} The robot picks up the left cup from the table and stacks it onto the right cup.
    \item \textbf{Stack the right cup onto the other cup.} The robot picks up the right cup from the table and stacks it onto the left cup.
    \item \textbf{Insert the straw into the cup.} The robot picks up a straw from the table and inserts it vertically into a cup.
    \item \textbf{Take the toast from the toaster and set it on a plate.} The robot picks the toast from the toaster and places it on a plate.
    \item \textbf{Unplug the charger.} The robot grasps the charger from the power strip, pulls out, and places on table.
\end{enumerate}

\myparagraph{Protocol and metrics.}
For each task, we design two distinct evaluation scenes with various distractors (Fig.~\ref{exp:fig:appdxB_test}), and evaluate each scene four times, resulting in eight trials per task.
Before each rollout, we reset the robot's initial pose and the object layout according to a pre-recorded mask to ensure a relatively fair comparison, and all evaluation scenes are different from the training data.
Since robot configurations differ physically from humans, achieving non-zero success on novel tasks with no real-robot data is challenging; we therefore report both the success rate and the average progress, where the latter follows the fine-grained, per-task scoring criteria in Fig.~\ref{exp:fig:appdxB_test} to quantify partial task completion.



\begin{figure}[t]
    \centering
    \includegraphics[width=\linewidth]{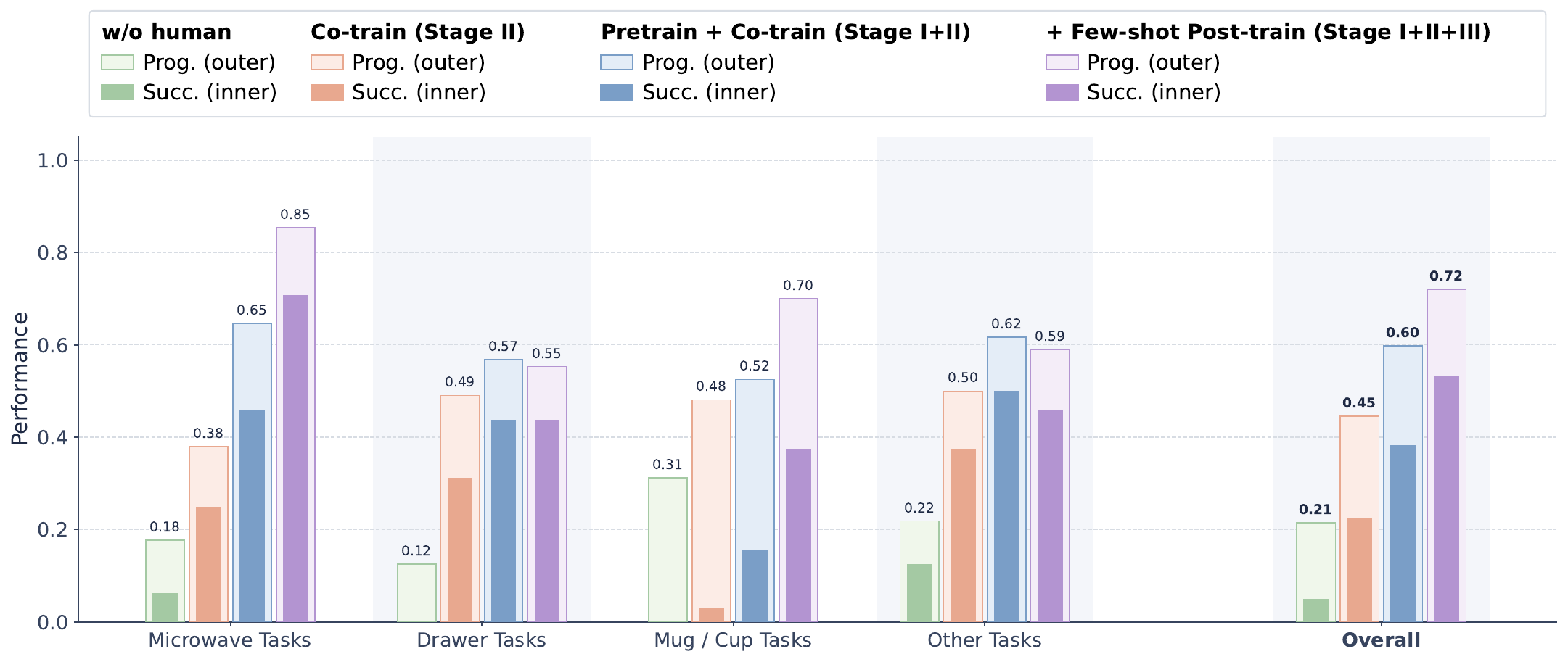}
    \caption{
        \textbf{Main results.}
        Training only on robot pick-and-place data (\textcolor[RGB]{140, 180, 150}{green}) is not enough for our downstream evaluation tasks.
        However, the co-training of human and robot actions (\textcolor[RGB]{232, 168, 143}{orange}) can efficiently transfer manipulation knowledge from human data to robot actions.
        The bridging action can also benefit from large-scale human-only pre-training (\textcolor[RGB]{129, 157, 195}{blue}), and from additional few-shot robot demonstrations (\textcolor[RGB]{200, 160, 220}{purple}).
    }
    \label{exp:fig:sec5.2-main}
\end{figure}

\subsection{The bridging action is scalable and essential for manipulation skill transfer}
\label{exp:sec:principle}

\begin{figure}[t]
    \centering
    \includegraphics[width=1\linewidth]{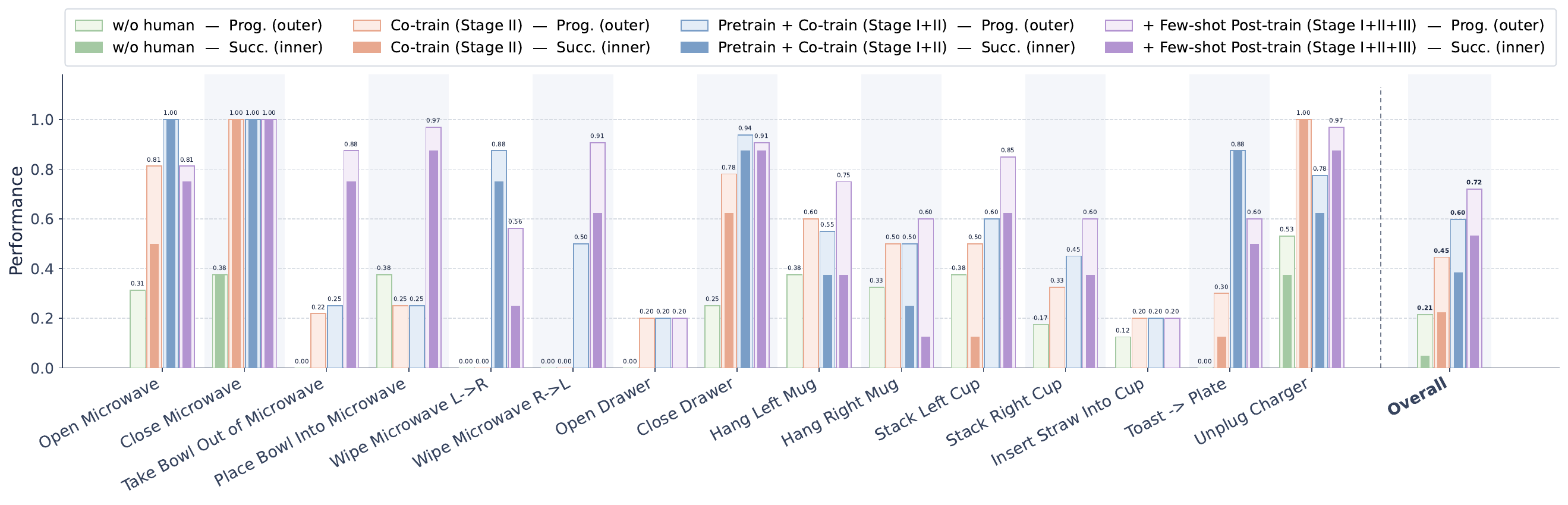}
    \caption{
        \textbf{Per-task comparison across training settings.}
        Training only on robot pick-and-place data (\textcolor[RGB]{140, 180, 150}{green}) is not enough for our downstream evaluation tasks.
        However, the co-training of human and robot actions (\textcolor[RGB]{232, 168, 143}{orange}) can efficiently transfer manipulation knowledge from human data to robot actions.
        The bridging action can also benefit from large-scale human-only pre-training (\textcolor[RGB]{129, 157, 195}{blue}), and from additional few-shot robot demonstrations (\textcolor[RGB]{200, 160, 220}{purple}).
    }
    \label{exp:fig:appdxB_main_results}
\end{figure}

\myparagraph{Finding 1: The bridging representation transfers skills beyond pick-and-place.}
An alternative hypothesis is that the robot could already solve the evaluation tasks simply by generalizing from large-scale pick-and-place data.
To rule this out, we compare two checkpoints trained on generalized robot pick-and-place data with (\textcolor[RGB]{232, 168, 143}{orange}) and without (\textcolor[RGB]{140, 180, 150}{green}) human data in Fig.~\ref{exp:fig:sec5.2-main}.
We also provide per-task performance in Fig.~\ref{exp:fig:appdxB_main_results}.
For better generalization, we co-trained both models on the same vision-language data.
The results show that training on pick-and-place data alone (\textcolor[RGB]{140, 180, 150}{green}) achieves very low performance on all tasks, showing that the robot is not able to solve the evaluation tasks only with the pick-and-place data.
We also observe that when co-training the model with human actions, there are substantial improvements on both task progress and success rate.
Such improvement also validates that our proposed bridging actions help to learn the human-to-robot motion transfer on manipulation tasks exceeding simple pick-and-place.

\myparagraph{Finding 2: The bridging representation scales with large-scale human-only pre-training.}
Human data is widely regarded as a highly scalable data source. 
After our preliminary verification that human data enables skill transfer via the bridging action, we aim to leverage large-scale human-only pre-training to further validate the scalability of this representation.
We compare human-robot co-training with (\textcolor[RGB]{129, 157, 195}{blue}) and without (\textcolor[RGB]{232, 168, 143}{orange}) human-only pre-training (Stage I).
As is shown in Fig.~\ref{exp:fig:appdxB_main_results}, though the policy is pre-trained with only the bridging action $\mathbf{a}^{\text{3D-wrist}}_{t}$, such large-scale human-only pre-training substantially improves the performance, confirming that the bridging action representation is scalable and effective.

\subsection{Why translation-only instead of 6DoF human wrist actions?}
\label{exp:sec:vs6dof}

\begin{figure}[htbp]
    \centering
    \includegraphics[width=\linewidth]{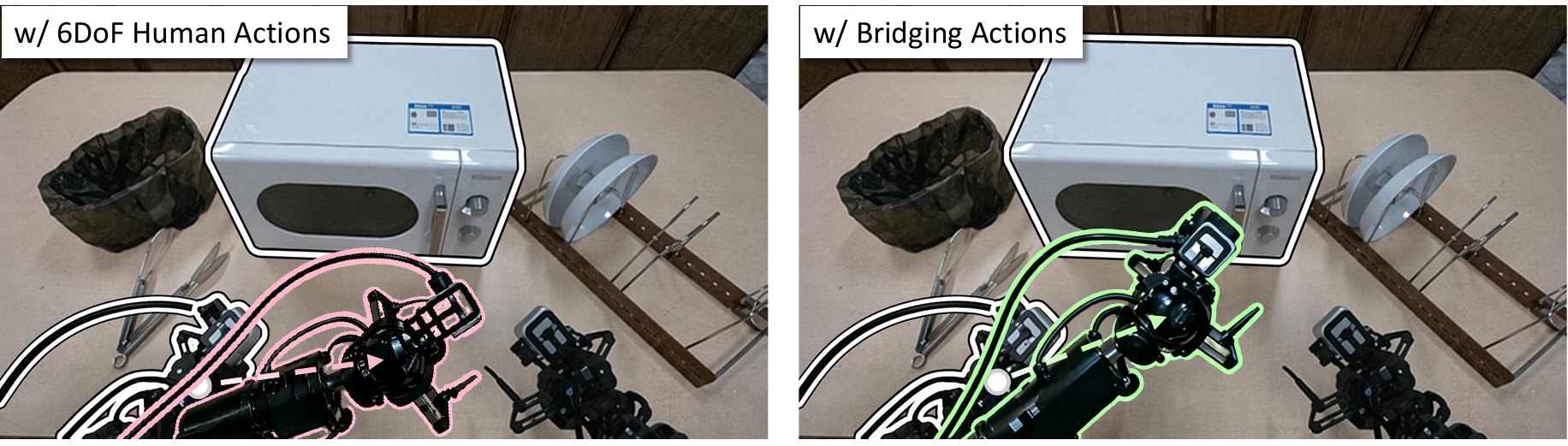}
    \caption{
        \textbf{Qualitative comparison on co-training with different human actions.}
        We show that when co-training on 6DoF human actions (left), the robot yields a distorted, off-target wrist pose; while our bridging actions (right) yield a natural pose that aligns with the microwave door handle.
    }
    \label{exp:fig:sec5.3_traj_frame}
\end{figure}

\begin{figure}[htbp]
    \centering
    \includegraphics[width=1\linewidth]{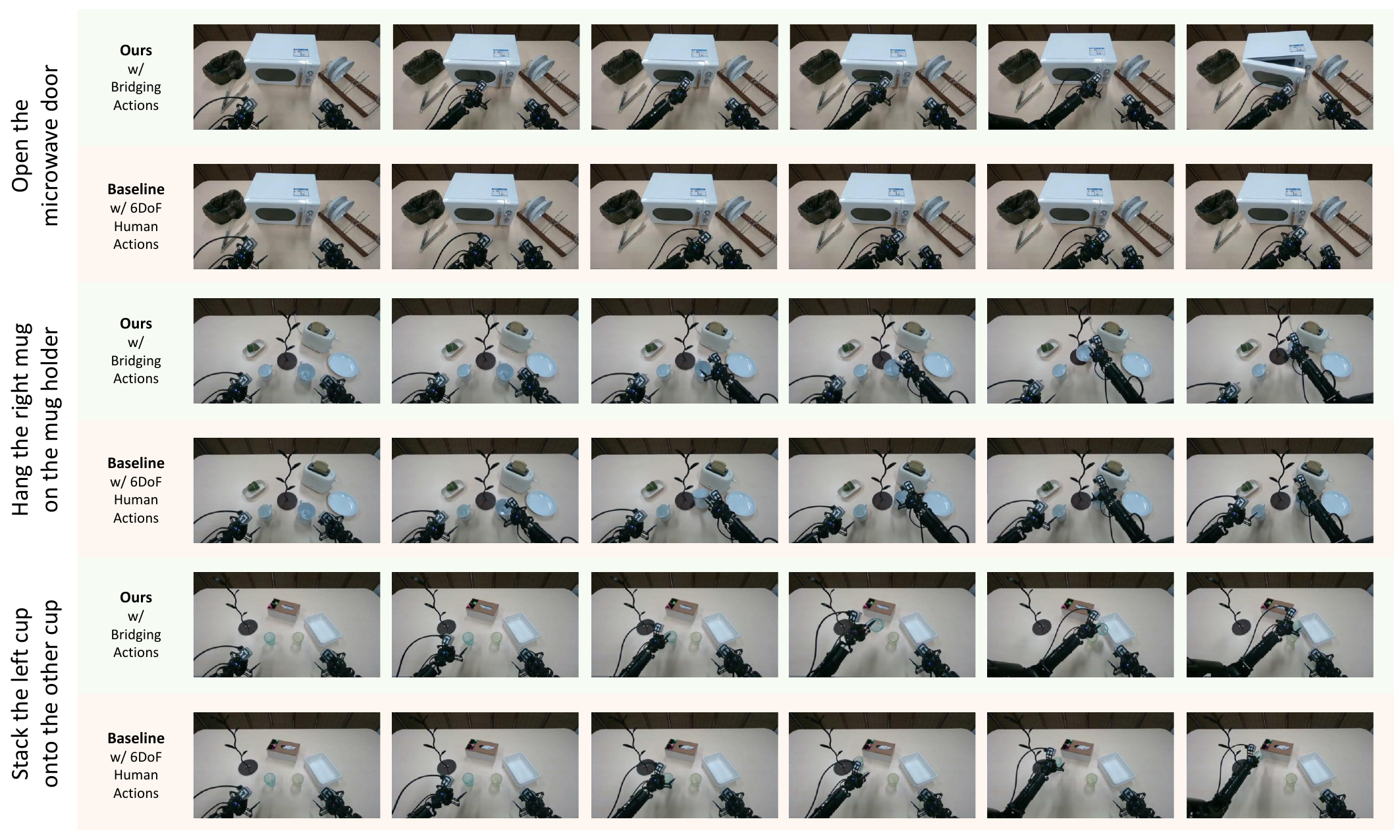}
    \caption{
        \textbf{Qualitative comparisons on co-training with different human actions.}
        Co-training with bridging actions (\textcolor[RGB]{152, 180, 134}{green}) enables more stable manipulation behaviors than the 6DoF baseline (\textcolor[RGB]{199, 159, 129}{orange}).
    }
    \label{exp:fig:appdxA_6DoF}
\end{figure}

The mainstream practice for human-robot co-training treats the human hand as another robotic embodiment by extracting relative 6DoF wrist actions~\cite{kareer2025egomimic,punamiya2025egobridge,zheng2026egoscale,cheang2025gr3}.
We compare this baseline against our bridging action $\mathbf{a}^{\text{3D-wrist}}_{t}$ in Tab.~\ref{exp:tab:sec5.3} by co-training the two models from scratch.
We also provide the qualitative analysis for the two models in Fig.~\ref{exp:fig:sec5.3_traj_frame} and Fig.~\ref{exp:fig:appdxA_6DoF}.
We observe that co-training with 6DoF human actions will lead to noisy and twisted behaviors, while co-training with our proposed bridging actions lead to more stable behaviors.
We also provide the qualitative comparisons in Tab.~\ref{exp:tab:sec5.3}, which also shows the superiority of the translation-only bridging action.

\begin{table}[htbp]
\centering
\caption{
    \textbf{Comparison on human action representation.}
    We compare 6DoF and translation-only (3DoF) human actions for human--robot co-training (from scratch).
    Our bridging action achieves better quantitative results on evaluation tasks.
}
\resizebox{\textwidth}{!}{
    \begin{tabular}{ccccccccccc}
    \toprule
    \multirow{2}{*}{Human Actions} & \multicolumn{2}{c}{Microwave Tasks} & \multicolumn{2}{c}{Drawer Tasks} & \multicolumn{2}{c}{Mug / Cup Tasks} & \multicolumn{2}{c}{Other Tasks} & \multicolumn{2}{c}{Overall} \\
    \cmidrule(lr){2-3} \cmidrule(lr){4-5} \cmidrule(lr){6-7} \cmidrule(lr){8-9} \cmidrule(lr){10-11}
     & Prog.(\%) & Succ.(\%) & Prog.(\%) & Succ.(\%) & Prog.(\%) & Succ.(\%) & Prog.(\%) & Succ.(\%) & Prog.(\%) & Succ.(\%) \\
    \midrule $\mathbf{a}^{\text{6D-eef}}$ & 25.00 & 4.17 & \textbf{55.00} & 31.25 & 28.13 & 0.00 & 49.17 & 33.33 & 34.67 & 12.50 \\
    $\mathbf{a}^{\text{3D-wrist}}$ & \textbf{38.02} & \textbf{25.00} & 49.06 & 31.25 & \textbf{48.13} & \textbf{3.13} & \textbf{50.00} & \textbf{37.50} & \textbf{44.58} & \textbf{22.50} \\\bottomrule
    \end{tabular}
}
\label{exp:tab:sec5.3}
\end{table}

\subsection{Human-only pre-training improves post-training data efficiency}
\label{exp:sec:few-shot}

From the above comparisons, we mainly put emphasis on the comparisons where no robot trajectories is involved for downstream evaluation tasks.
In this section, we show that the bridging actions can improve post-training efficiency on few-shot robot trajectories.

We compare the few-shot post-training (10 robot trajectories per task) with and without human-only pre-training in Tab.~\ref{exp:tab:sec5.4}.
During human-only pre-training (Stage I), we only supervise the model to predict the bridging signal $\mathcal{L}_{\text{FM}}^{\text{3D-wrist}}$, while including all action components for post-training (Stage III).
We show that though the model is not exposed to executable actions during human-only pre-training, it still substantially improves the average progress and success rate during post-training.
This suggests that our model can benefit from the pre-training knowledge and achieve higher post-training efficiency.

\begin{table}[htbp]
\centering
\caption{
    \textbf{Post-training data efficiency comparison after human-only pre-training.}
    Though the model is only exposed to non-executable human actions $\mathbf{a}_{t}^{\text{3D-wrist}}$ during pre-training, we can efficiently improve the data efficiency for few-shot robot post-training.
}
\resizebox{\textwidth}{!}{
    \begin{tabular}{ccccccccccc}
    \toprule
    \multirow{2}{*}{Model} & \multicolumn{2}{c}{Microwave Tasks} & \multicolumn{2}{c}{Drawer Tasks} & \multicolumn{2}{c}{Mug / Cup Tasks} & \multicolumn{2}{c}{Other Tasks} & \multicolumn{2}{c}{Overall} \\
    \cmidrule(lr){2-3} \cmidrule(lr){4-5} \cmidrule(lr){6-7} \cmidrule(lr){8-9} \cmidrule(lr){10-11}
     & Prog.(\%) & Succ.(\%) & Prog.(\%) & Succ.(\%) & Prog.(\%) & Succ.(\%) & Prog.(\%) & Succ.(\%) & Prog.(\%) & Succ.(\%) \\
    \midrule
    Stage III & 71.77 & 58.33 & \textbf{56.88} & \textbf{43.75} & 37.50 &  6.25 & 37.50 & 25.00 & 53.79 & 35.83 \\
    Stage I + III & \textbf{80.73} & \textbf{68.75} & 44.38 & 25.00 & \textbf{71.25} & \textbf{46.88} & \textbf{70.00} & \textbf{58.33} & \textbf{71.21} & \textbf{55.00} \\
    \bottomrule
    \end{tabular}
}
\label{exp:tab:sec5.4}
\end{table}

\subsection{Training bridging actions for robot data is essential for manipulation transfer}
\label{sec:exp:bridging_tasks}

During human-robot co-training, we randomly add the bridging action $\mathbf{a}^{\text{3D-wrist}}$ to, or substitute it for, $\mathbf{a}^{\text{6D-eef}}$ as the prediction objective for robot data (Sec.~\ref{sec:method:strategy}).
To validate this design, we compare two checkpoints co-trained on human and robot actions with and without this augmentation, both built upon the large-scale human-only pre-training.
As shown in Tab.~\ref{exp:tab:sec5.5}, removing the bridging training objective greatly degrades the performance across all task groups, dropping the overall success rate from $38.33\%$ to $12.50\%$.

\begin{table}[htbp]
\centering
\caption{
    \textbf{Ablation of the bridging objective on robot data.}
    During co-training, it is essential to also supervise the bridging action on robot data for manipulation skill transfer.
}
\resizebox{\textwidth}{!}{
    \begin{tabular}{ccccccccccc}
    \toprule
    \multirow{2}{*}{Robot Actions} & \multicolumn{2}{c}{Microwave Tasks} & \multicolumn{2}{c}{Drawer Tasks} & \multicolumn{2}{c}{Mug / Cup Tasks} & \multicolumn{2}{c}{Other Tasks} & \multicolumn{2}{c}{Overall} \\
    \cmidrule(lr){2-3} \cmidrule(lr){4-5} \cmidrule(lr){6-7} \cmidrule(lr){8-9} \cmidrule(lr){10-11}
     & Prog.(\%) & Succ.(\%) & Prog.(\%) & Succ.(\%) & Prog.(\%) & Succ.(\%) & Prog.(\%) & Succ.(\%) & Prog.(\%) & Succ.(\%) \\
    \midrule
    w/o $\mathbf{a}^{\text{3D-wrist}}$  & 35.73 & 10.42 & 39.38 & 12.50 & 39.38 & 0.00 & 48.13 & 33.33 & 39.67 & 12.50 \\
    w/ $\mathbf{a}^{\text{3D-wrist}}$ & \textbf{64.58} & \textbf{45.83} & \textbf{56.88} & \textbf{43.75} & \textbf{52.50} & \textbf{15.63} & \textbf{61.67} & \textbf{50.00} & \textbf{59.75} & \textbf{38.33} \\
    \bottomrule
    \end{tabular}
}
\label{exp:tab:sec5.5}
\end{table}

\subsection{Human-only pre-training aligns with the executable robot action space}
\label{exp:sec:alignment}

\begin{figure}[h]
    \centering
    \includegraphics[width=0.85\linewidth]{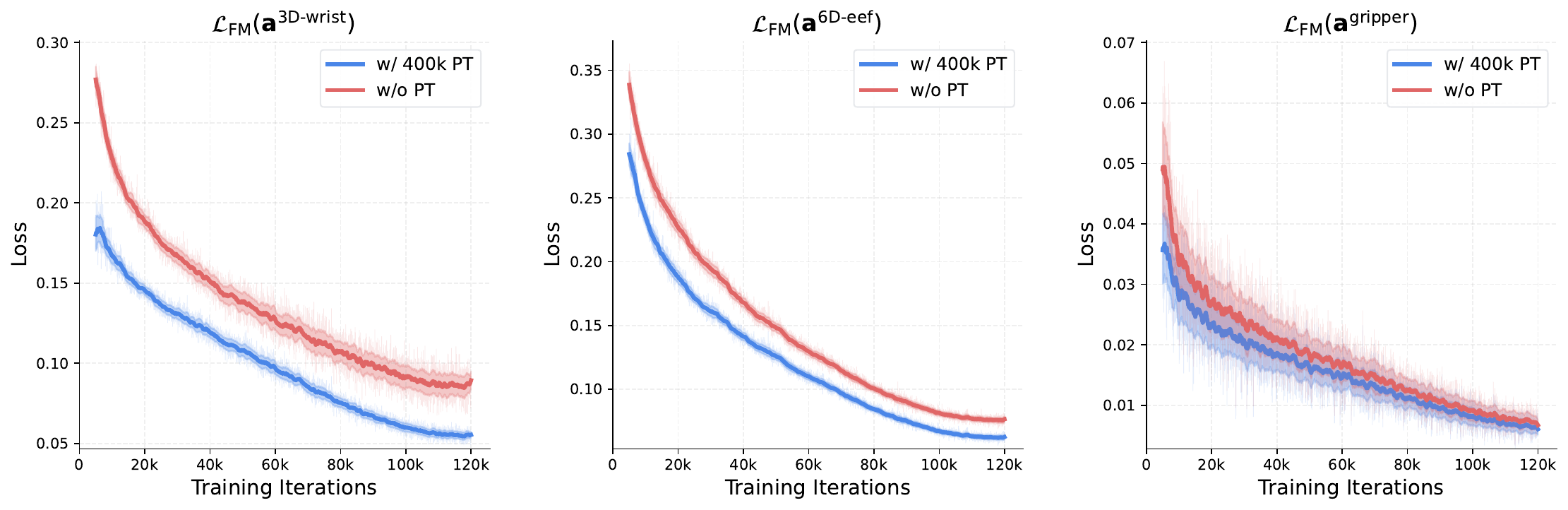}
    \caption{
        \textbf{Training loss comparison during human-robot co-training.}
        Although human-only pre-training supervises only wrist translations, it accelerates the convergence of the other action components, \textit{i.e.}, the 6DoF end-effector action and the gripper action.
    }
    \label{exp:fig:pretraining_curve}
\end{figure}

Previously, we investigated whether human-only pre-training improves model performance on real-world robotic evaluation tasks in Sec.~\ref{exp:sec:principle}.
As a complementary analysis, we further present the training loss of each action component during co-training in Fig.~\ref{exp:fig:pretraining_curve}. 
This allows us to examine whether the training objective for our bridging actions aligns with that of 6DoF end-effector actions.
Specifically, we compare two experiments, one initialized from scratch (\textcolor[RGB]{224, 102, 102}{red}), and another initialized from human-only pre-training (\textcolor[RGB]{74, 134, 232}{blue}).
We find that though human-only pre-training supervises only the non-executable wrist-translation action, it yields a lower training loss for both the 6DoF end-effector action and the gripper action during co-training.
This implies that optimizing the bridging signal $\mathbf{a}^{\text{3D-wrist}}$ shares a similar objective landscape with the executable $\mathbf{a}^{\text{6D-eef}}$, explaining why the translation-only pre-training transfers to the full robot action space.
This loss-level alignment is also consistent with the held-out gains reported earlier: human-only pre-training improves downstream task success and progress both in human-robot co-training in Sec.~\ref{exp:sec:principle}.

\subsection{The bridging action aligns with the executable robot actions}
\label{exp:sec:alignment2}

\begin{figure}[htbp]
    \centering
    \includegraphics[width=1\linewidth]{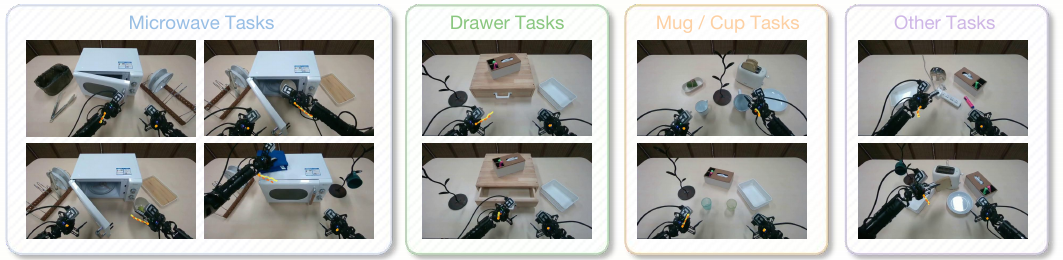}
    \caption{
        \textbf{Visualization of the bridging action and 6DoF end-effector action prediction.}
        Base on the same vision and language input, we ask the model to produce bridging actions and end-effector actions, and then project both actions onto the head camera.
        The two types of actions align closely across diverse tasks.
    }
    \label{exp:fig:wrist_eef_alignment}
\end{figure}

Another critical question is whether the predicted bridging action $\mathbf{a}^{\text{3D-wrist}}$ is consistent with the 6DoF end-effector action $\mathbf{a}^{\text{6D-eef}}$ for the robot.
To answer this question, we ask the model to produce bridging actions and end-effector actions base on the same vision and language input, and then project both actions onto the head camera based on the robot's state in Fig.~\ref{exp:fig:wrist_eef_alignment}.
The generated $\mathbf{a}^{\text{3D-wrist}}$ is shown in \textcolor[RGB]{207, 200, 92}{yellow} and $\mathbf{a}^{\text{6D-eef}}$ in \textcolor[RGB]{191, 88, 36}{orange}.
We observe that the two types of actions align closely across diverse tasks for both arms, indicating that $\mathbf{a}^{\text{3D-wrist}}$ can serve as a reliable alternative for real-robot actions when learning transferable manipulation knowledge.

\subsection{The upper bound of the bridging objective}
\label{exp:sec:upperbound}
\begin{figure}[htbp]
    \centering
    \includegraphics[width=1\linewidth]{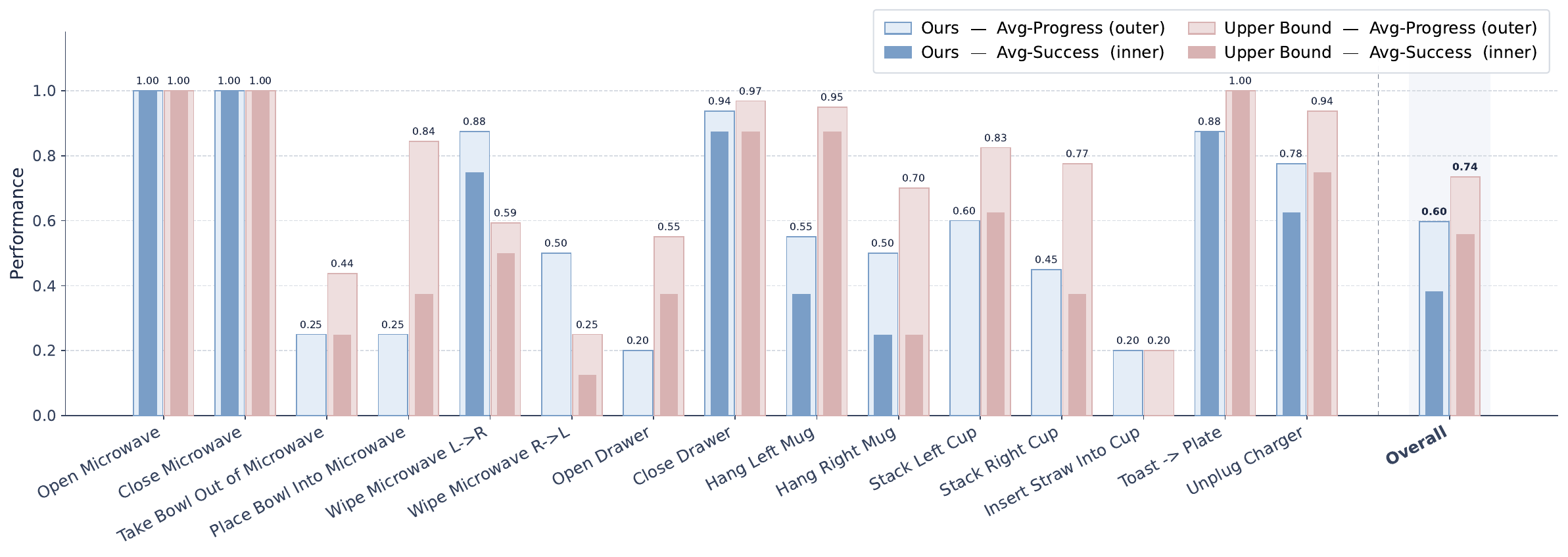}
    \caption{
        \textbf{Per-task comparison for the upper-bound analysis.}
        Applying the same training objective as human actions to task-specific in-lab robot actions, the manipulation knowledge transfer becomes increasingly efficient as the visual gap and action noise diminish.
    }
    \label{exp:fig:appdxB_upper_bound}
\end{figure}

To probe the performance upper bound of the bridging representation, we remove the human-robot gap by treating \textbf{task-specific robot demonstrations} (100 traj/task) as if they were human action data.
Specifically, we transform them into translation-only wrist actions $\mathbf{a}^{\text{3D-wrist}}$ and apply the same training objective as for human actions, but with no observation gap (wrist cameras are available) and far less action noise.
As reported in Tab.~\ref{exp:tab:sec5.6} and detailed per-task in Fig.~\ref{exp:fig:appdxB_upper_bound}, this upper-bound variant (\textcolor[RGB]{216, 178, 178}{brown}) substantially outperforms the default co-training with human data (\textcolor[RGB]{129, 157, 195}{blue}).
This confirms that 1) the bridging representation itself is an effective medium for skill transfer, and 2) the transfer becomes increasingly efficient as the visual gap and action noise diminish, suggesting a promising direction for broader multi-embodiment learning.

\begin{table}[htbp]
\centering
\caption{
    \textbf{Upper bound of the bridging objective.}
    Applying the same training objective as human actions to task-specific in-lab robot actions, the bridging action itself enables even stronger skill transfer when there is no embodiment gap.
}
\resizebox{\textwidth}{!}{
    \begin{tabular}{ccccccccccc}
    \toprule
    \multirow{2}{*}{Model} & \multicolumn{2}{c}{Microwave Tasks} & \multicolumn{2}{c}{Drawer Tasks} & \multicolumn{2}{c}{Mug / Cup Tasks} & \multicolumn{2}{c}{Other Tasks} & \multicolumn{2}{c}{Overall} \\
    \cmidrule(lr){2-3} \cmidrule(lr){4-5} \cmidrule(lr){6-7} \cmidrule(lr){8-9} \cmidrule(lr){10-11}
     & Prog.(\%) & Succ.(\%) & Prog.(\%) & Succ.(\%) & Prog.(\%) & Succ.(\%) & Prog.(\%) & Succ.(\%) & Prog.(\%) & Succ.(\%) \\
    \midrule
    Default (Ours) & 64.58 & 45.83 & 56.88 & 43.75 & 52.50 & 15.63 & 61.67 & 50.00 & 59.75 & 38.33 \\
    \textbf{Upper Bound} & \textbf{68.75} & \textbf{54.17} & \textbf{75.94} & \textbf{62.50} & \textbf{81.25} & \textbf{53.13} & \textbf{71.25} & \textbf{58.33} & \textbf{73.54} & \textbf{55.83} \\
    \bottomrule
    \end{tabular}
}
\label{exp:tab:sec5.6}
\end{table}

\subsection{Failure Case Analysis}
\label{exp:sec:failure}

\begin{figure}[htbp]
    \centering
    \includegraphics[width=1\linewidth]{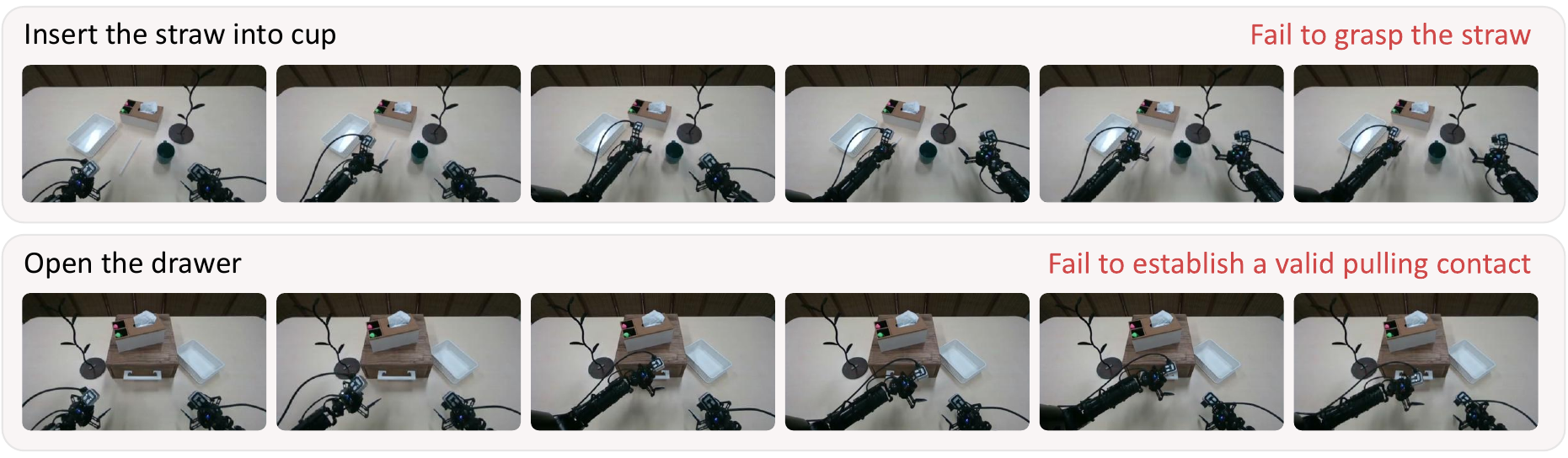}
    \caption{
        \textbf{Failure case analysis.}
        We visualize representative failure trajectories for ``\texttt{insert the straw into the cup}'' and ``\texttt{open the drawer}''.
        Although the robot demonstrates clear task intent, it often fails at critical moments, such as grasping the straw securely or rotating the wrist appropriately to establish a valid pulling contact.
    }
    \label{exp:fig:appdxB_case}
\end{figure}

Failure cases mainly arise in tasks where success depends not only on reaching the correct interaction region, but also on a precise end-effector configuration during contact-rich manipulation.
The most representative examples are ``\texttt{insert the straw into the cup}'' and ``\texttt{open the drawer}'', where the policy often demonstrates clear task intent yet fails at critical steps, such as grasping the straw securely or rotating the wrist to establish a valid pulling contact, as shown in Fig.~\ref{exp:fig:appdxB_case}.
These failures are consistent with our design choice of discarding rotational supervision from human data, and point to incorporating limited, reliable rotation cues as a promising direction for future work.

%% file: sections/conclusion.tex
\section{Conclusions}

\myparagraph{Limitations and Future Work.} 
To leverage in-the-wild human actions, we adopt only wrist translation as the bridging action, which limits transfer on tasks that require fine rotational adjustments (Sec.~\ref{exp:sec:failure}). 
We also find the robot struggles to pick up thin objects after co-training, which we attribute mainly to 1) the observation and embodiment gap and 2) the inevitable noise in human actions.
In future work, with larger-scale and more diverse robot actions, we can further narrow the gap between human and robot actions.

\myparagraph{Conclusions.}
In this work, we investigate the feasibility of learning task-specific bi-manual robot skills from human motion data.
We propose a translation-based bridging action representation that is compatible with both robot and human action data. 
To address the potentially missing action components in different data sources, we further introduce an interleaved action sequence representation.
Experimental results show that the proposed translation-only actions can efficiently transfer human manipulation knowledge to robot skill learning and effectively benefit from large-scale pretraining.